\documentclass[sigconf, nonacm]{acmart}
\usepackage{supertabular}
\usepackage{longtable}
\usepackage{booktabs}
\usepackage{array}
\usepackage{siunitx}
\usepackage{booktabs}
\usepackage{tabularx}
\usepackage{siunitx}
\usepackage{graphicx}
\usepackage[section]{placeins}

\usepackage{tabularray}
\UseTblrLibrary{booktabs,siunitx} 
\newcolumntype{L}[1]{>{\raggedright\arraybackslash}p{#1}}

\newcommand\vldbdoi{XX.XX/XXX.XX}
\newcommand\vldbpages{XXX-XXX}
\newcommand\vldbvolume{14}
\newcommand\vldbissue{1}
\newcommand\vldbyear{2020}
\newcommand\vldbauthors{\authors}
\newcommand\vldbtitle{\shorttitle} 
\newcommand\vldbavailabilityurl{https://github.com/LLwork8888/CreditAudit}
\newcommand\vldbpagestyle{plain} 

\begin{document}
\title{CreditAudit: 2$^\text{nd}$ Dimension for LLM Evaluation and Selection}

\settopmatter{authorsperrow=3}

\newcommand{\AffNPUTeleAI}{%
  \affiliation{%
    \institution{Northwestern Polytechnical University\\ Institute of Artificial Intelligence (TeleAI), China Telecom}
    \country{China}
  }%
}
\newcommand{\AffDMUTeleAI}{%
  \affiliation{%
    \institution{Dalian Maritime University\\ Institute of Artificial Intelligence (TeleAI), China Telecom}
    \country{China}
  }%
}
\newcommand{\AffTeleAI}{%
  \affiliation{%
    \institution{Institute of Artificial Intelligence (TeleAI), China Telecom}
    \country{China}
  }%
}
\newcommand{\AffGXNUTeleAI}{%
  \affiliation{%
    \institution{Guangxi Normal University\\ Institute of Artificial Intelligence (TeleAI), China Telecom}
    \country{China}
  }%
}

\author{Yiliang~Song}
\authornote{Work done at TeleAI.}
\authornote{These authors contributed equally to this work.}
\AffGXNUTeleAI

\author{Hongjun~An}
\authornotemark[1]
\authornotemark[2]
\AffNPUTeleAI

\author{Jiangong~Xiao}
\authornotemark[1]
\AffNPUTeleAI

\author{Haofei~Zhao}
\authornotemark[1]
\AffNPUTeleAI

\author{Jiawei~Shao}
\AffTeleAI

\author{Xuelong~Li}
\authornote{Corresponding author.}
\email{xuelong\_li@ieee.org}
\AffTeleAI

\renewcommand{\shortauthors}{Song et al.}

\begin{teaserfigure}
  \includegraphics[trim={0cm 0.25cm 0cm 0.0cm}, clip, width=\textwidth]{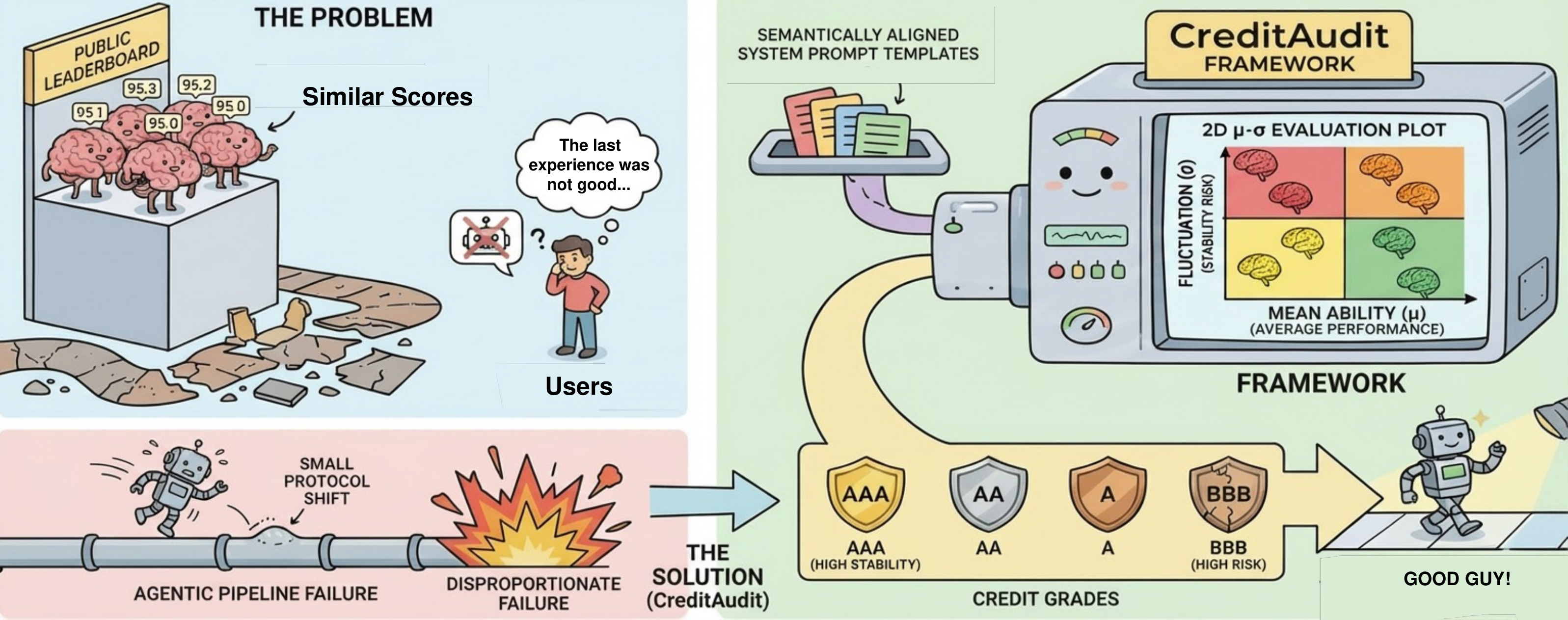}
\caption{CreditAudit audits LLM stability under semantically aligned system-prompt variants, reporting mean ability $\mu$, fluctuation risk $\sigma$, and credit grades for tiered deployment.}
  \label{fig:teaser}
\end{teaserfigure}

\begin{abstract}
Leaderboard scores on public benchmarks have been steadily rising and converging, with many frontier language models now separated by only marginal differences.
However, these scores often fail to match users' day to day experience, because system prompts, output protocols, and interaction modes evolve under routine iteration, and in agentic multi step pipelines small protocol shifts can trigger disproportionate failures, leaving practitioners uncertain about which model to deploy.
We propose CreditAudit, a deployment oriented credit audit framework that evaluates models under a family of semantically aligned and non adversarial system prompt templates across multiple benchmarks, reporting mean ability as average performance across scenarios and scenario induced fluctuation sigma as a stability risk signal, and further mapping volatility into interpretable credit grades from AAA to BBB via cross model quantiles with diagnostics that mitigate template difficulty drift.
Controlled experiments on GPQA, TruthfulQA, and MMLU Pro show that models with similar mean ability can exhibit substantially different fluctuation, and stability risk can overturn prioritization decisions in agentic or high failure cost regimes.
By providing a 2D and grade based language for regime specific selection, CreditAudit supports tiered deployment and more disciplined allocation of testing and monitoring effort, enabling more objective and trustworthy model evaluation for real world use.
\end{abstract}

\maketitle

\pagestyle{\vldbpagestyle}
\begingroup\small\noindent\raggedright\textbf{PVLDB Reference Format:}\\
\vldbauthors. \vldbtitle. PVLDB, \vldbvolume(\vldbissue): \vldbpages, \vldbyear.\\
\href{https://doi.org/\vldbdoi}{doi:\vldbdoi}
\endgroup
\begingroup
\renewcommand\thefootnote{}\footnote{\noindent
This work is licensed under the Creative Commons BY-NC-ND 4.0 International License. Visit \url{https://creativecommons.org/licenses/by-nc-nd/4.0/} to view a copy of this license. For any use beyond those covered by this license, obtain permission by emailing \href{mailto:info@vldb.org}{info@vldb.org}. Copyright is held by the owner/author(s). Publication rights licensed to the VLDB Endowment. \\
\raggedright Proceedings of the VLDB Endowment, Vol. \vldbvolume, No. \vldbissue\ %
ISSN 2150-8097. \\
\href{https://doi.org/\vldbdoi}{doi:\vldbdoi} \\
}\addtocounter{footnote}{-1}\endgroup

\ifdefempty{\vldbavailabilityurl}{}{
\vspace{.3cm}
\begingroup\small\noindent\raggedright\textbf{PVLDB Artifact Availability:}\\
The source code, data, and/or other artifacts have been made available at \url{\vldbavailabilityurl}.
\endgroup
}

\section{Introduction}
\label{sec:intro}

Public leaderboards show continued score gains alongside cross vendor convergence, with leading LLMs separated by ever smaller margins\cite{opencompass2023, openllmleaderboard2023}. 
Yet practitioners and end users often experience meaningful gaps in reliability, especially when system prompts, output protocols, and interaction modes change during routine product iteration. 
This creates a puzzling selection landscape where leaderboard rank feels less predictive than expected. 
The situation resembles financial credit auditing, where an Asset Pricing is not judged by average performance in one market condition but by how stably it behaves under normal regime shifts, because operational risk is often driven by volatility and tail events rather than by the mean.

To make benchmark scores more objective, prior work has mainly pursued breadth, scale, and methodological discipline \cite{chang2024survey}. 
For breadth, evaluation has expanded from single-task scoring to multi-capability coverage across knowledge, reasoning, math, and code, via broad suites and specialized benchmarks \cite{hendrycks2021mmlu,srivastava2022bigbench,liang2022helm,cobbe2021gsm8k,chen2021humaneval,austin2021mbpp,zheng2023judging,chiang2024arena}. 
For scale and headroom, benchmarks have been revised or strengthened to remain discriminative as frontier models improve, including harder task sets and refined question construction \cite{suzgun2022bbh,rein2023gpqa,wang2024mmlupro,myrzakhan2024openllm}. 
Within these regimes, a growing line of sensitivity studies shows that scores and even rankings can shift under small and seemingly benign protocol choices, such as instruction paraphrases, formatting conventions, and multiple-choice option ordering, motivating multi-prompt evaluation and distributional reporting rather than single-template summaries \cite{mizrahi2024state,alzahrani2024benchmarks}. 
Complementary reproducibility and uncertainty work argues for controlling evaluation artifacts and quantifying variability induced by stochasticity and experimental repeats \cite{biderman2024lessons,blackwell2024uncertainty}. 
Despite these advances, a gap remains between evaluation methodology and selection decisions: while prompt sensitivity and uncertainty are increasingly documented \cite{mizrahi2024state,blackwell2024uncertainty}, it is still relatively uncommon for mainstream benchmarking and leaderboard practice to (i) systematically sample a structured family of semantically aligned and non-adversarial \emph{system} prompts, (ii) summarize protocol-induced variation as a first-class, cross-model comparable risk statistic alongside mean performance, and (iii) translate that risk signal into a decision-ready tier that directly supports regime-specific model prioritization \cite{chang2024survey,nitsure2023risk,myrzakhan2024openllm}.

Motivated by these gaps, we propose \textbf{CreditAudit}, a selection-oriented protocol-robustness audit for LLM evaluation and prioritization. 
CreditAudit addresses the first gap by systematically sampling a structured family of semantically aligned and non-adversarial system-prompt templates that preserve question semantics while perturbing interaction protocol in realistic ways, and by applying the same template indices consistently across benchmarks to enable controlled cross-scenario comparison. 
It addresses the second gap by summarizing each model with a two-dimensional audit outcome: mean ability $\mu$, computed as the average accuracy across scenarios and benchmarks, and scenario-induced fluctuation $\sigma$, computed as the standard deviation across scenarios, which we interpret as a protocol-sensitivity risk signal under benign variation. 
It addresses the third gap by converting $\sigma$ into an interpretable credit grade from AAA to BBB using cross-model quantiles, producing a decision-ready risk tier that supports regime-specific prioritization and tiered selection. 
To make the risk signal trustworthy, CreditAudit further provides diagnostics designed to reduce template difficulty-drift artifacts and to attribute observed volatility to model sensitivity rather than to globally harder or easier templates. 
Finally, we release an open-source CreditAudit toolkit that implements scenario construction, subset sampling, multi-model evaluation runners, aggregation of $(\mu,\sigma)$, grade assignment, and automated reporting artifacts, enabling reproducible audits and easy extension to new models and benchmarks.

We evaluate CreditAudit on three complementary multiple-choice benchmarks, GPQA, TruthfulQA, and MMLU-Pro, using fixed shared evaluation subsets and aligned template indices for strict horizontal comparability. 
Across all evaluated models, we find that similar mean scores can mask substantial differences in robustness, with overall scenario-induced fluctuations spanning from $0.63$ to $2.63$ and corresponding credit grades ranging from AAA to BBB. 
Notably, robustness can overturn prioritization decisions that would appear reasonable under a single-score view, particularly in settings where protocol switching is frequent or failure costs are high. 
These results support a practical selection rule suggested by CreditAudit: prioritize low $\sigma$ as a first-order constraint, then optimize $\mu$ within an acceptable robustness tier, yielding a disciplined basis for model selection beyond leaderboard scores.

Our contributions can be summarized as follows:
\begin{itemize}
    \item We propose \textbf{CreditAudit}, a selection-oriented protocol-robustness audit that evaluates LLMs under a structured family of semantically aligned, non-adversarial system prompts to expose performance variability under benign protocol shifts.
    \item We introduce a 2D audit outcome consisting of mean ability $\mu$ and scenario-induced fluctuation $\sigma$, and map $\sigma$ to interpretable credit grades via cross-model quantiles, yielding a decision-ready risk tier for regime-specific model prioritization.
    \item We design audit controls and diagnostics to reduce template difficulty drift and improve attribution of volatility to model sensitivity, including aligned template indexing across benchmarks and strict horizontal comparability via shared evaluation subsets.
    \item We release an open-source toolkit that implements scenario construction, evaluation, aggregation of $(\mu,\sigma)$, grade assignment, and automated reporting, enabling reproducible audits and easy extension to new models and benchmarks.
\end{itemize}

\section{Related Work}
\subsection{Capability Benchmarking and Public Leaderboards}
The evaluation landscape for large language models has rapidly expanded, spanning both benchmark-centric testing and public leaderboards that provide continuous, comparative signals for model capability \cite{chang2024survey,an2025ai,yuan2026informationcapacityevaluatingefficiency}. 
On the benchmark side, widely used multiple-choice and task-specific datasets aim to probe broad knowledge and reasoning \cite{hendrycks2021mmlu}, mathematical problem solving \cite{cobbe2021gsm8k}, and code generation \cite{chen2021humaneval,austin2021mbpp}. 
To improve coverage beyond any single dataset, broad evaluation suites and holistic frameworks further expand the capability surface and report multiple metrics and scenarios, reducing reliance on one benchmark as a proxy for overall quality \cite{srivastava2022bigbench,liang2022helm}. 
As frontier models improve, benchmarks have also been revised or strengthened to preserve headroom and discriminability, including harder task sets and refined question construction \cite{suzgun2022bbh,rein2023gpqa,wang2024mmlupro}. 

In parallel, public leaderboards and platforms have emerged as an operational layer that aggregates results, standardizes evaluation pipelines, and makes comparisons accessible. 
Representative examples include OpenCompass as a general evaluation platform \cite{opencompass2023}, the Open LLM Leaderboard and its updated variants as widely used benchmark aggregators \cite{openllmleaderboard2023,myrzakhan2024openllm}, and preference-based arenas such as Chatbot Arena and judge-based pipelines such as MT-Bench that approximate human preference at scale \cite{chiang2024arena,zheng2023judging}. 
These resources have been invaluable for tracking progress and broadening the set of measured capabilities, but they typically summarize models through a small number of prompt settings or a single reported score per benchmark, leaving limited visibility into how reliably a model behaves when interaction protocols shift during routine usage.

\subsection{Sensitivity, Uncertainty, and Risk-Aware Model Selection}
A complementary line of work argues that evaluation should account for sensitivity and uncertainty, as small protocol perturbations can induce meaningful score and ranking changes. 
Multi-prompt evaluation has been advocated as a response to prompt dependence, emphasizing that single-template reporting can misrepresent model quality \cite{mizrahi2024state}. 
Related efforts study prompt sensitivity more explicitly and provide frameworks for measuring performance variation under prompt changes \cite{alzahrani2024benchmarks}. 
In multiple-choice settings, sensitivity can arise from seemingly innocuous formatting decisions such as the order of answer options, or even minimal output constraints that change extraction and parsing behavior \cite{pezeshkpour2023optionorder}. 
PromptEval formalizes multi-prompt evaluation as a distributional estimation problem, enabling more efficient characterization of performance across prompt variants \cite{maia2024efficient}. 
Beyond prompt dependence, reproducibility and uncertainty analyses highlight that evaluation outcomes can vary due to stochasticity and experimental repeats, motivating tighter control of evaluation artifacts and explicit uncertainty reporting \cite{biderman2024lessons,blackwell2024uncertainty}. 
Finally, risk-aware benchmarking frameworks begin to frame model comparison as a decision problem under uncertainty, emphasizing that mean performance alone may be insufficient for selection \cite{nitsure2023risk}. 

Our work builds on these insights but targets a distinct decision interface. 
Rather than only documenting sensitivity or reporting uncertainty in isolation, CreditAudit operationalizes benign protocol variation through a structured family of semantically aligned \emph{system} prompts, summarizes the resulting variation as a comparable risk signal alongside mean ability, and converts that signal into an interpretable tier for regime-specific prioritization, with diagnostics designed to reduce template-drift artifacts and support actionable model selection.

\section{Preliminaries}
\label{sec:method}

\subsection{What LLM Evaluation Estimates: Partial Observability and Prompt-Induced Reliability}
Modern leaderboards provide a useful reference for model capability. Yet as score gaps shrink and ``benchmark chasing'' becomes common, practitioners often observe that leaderboard ranks do not consistently match real usage experience: models with similar average scores can behave very differently once the interaction setup changes slightly (e.g., system instruction wording, output-format constraints, or the degree of caution requested).
This mismatch is a natural consequence of how LLM evaluation differs from standard supervised learning evaluation in two fundamental aspects.

\paragraph{\textbf{The pretraining distribution is not fully observable.}}
For large pretrained models, the support of the pretraining corpus distribution $\mathcal{P}_{\mathrm{pre}}$ cannot be enumerated.
Hence, a correct answer on a benchmark item is generally not identifiable as coming from \emph{generalization} versus \emph{memorization}.
At the item level, correctness can be abstracted as an unidentifiable mixture:
\begin{equation}
\begin{aligned}
c_m(x, y; \pi) &= \alpha_x \, c_m^{\text{mem}}(x, y; \pi) \\
&\quad + (1 - \alpha_x) \, c_m^{\text{gen}}(x, y; \pi), \\
\alpha_x &\in [0, 1] \text{ unobserved}.
\end{aligned}
\label{eq:mix}
\end{equation}

where $m$ is a model, $(x,y)$ denotes an input question and its reference answer (or decision criterion), and $\pi$ denotes an interaction protocol.
Therefore, benchmark scores are best interpreted as evidence that a model \emph{exhibits} the capability under the evaluated conditions, rather than a clean proof that the capability is obtained via generalization.

\paragraph{\textbf{The interaction space is effectively unbounded.}}
LLM behavior depends not only on task content but also on the interaction protocol (system prompts, formatting constraints, response style requirements, etc.).
Such protocol variations are ubiquitous in real systems and are often ``benign'' rather than adversarial.
Treating protocol as part of the evaluation environment is essential for model selection, because performance can vary materially under normal protocol changes.

\subsection{Capability Under an Open Interaction Protocol Space}
We view an evaluation outcome as jointly determined by task content and the interaction protocol.
Let $(x,y)$ denote a question and its reference answer (or decision criterion), and let $\pi\in\Pi$ denote an interaction protocol, which may include the system prompt, output-format constraints, style requirements, and any instruction context that can shape model behavior.
Given $(x,\pi)$, model $m$ produces a free-form textual response
$r \sim p_m(\cdot \mid x,\pi)$.
Because the output is free text, we define a scoring function $h(r,y)\in[0,1]$ that maps a response to a degree of correctness, yielding the protocol-conditional expected correctness:
\begin{equation}
c_m(x,y;\pi)
=
\mathbb{E}_{r\sim p_m(\cdot\mid x,\pi)}\!\big[h(r,y)\big],
\qquad
h(r,y)\in[0,1].
\label{eq:cap_protocol}
\end{equation}

For multiple-choice evaluation with parsing, Eq.~\eqref{eq:cap_protocol} admits an equivalent conditional-probability interpretation.
Let $\hat{y}=g(r)$ be the parsed option from response $r$ under a fixed parsing rule $g(\cdot)$.
Then
\begin{equation}
c_m(x,y;\pi)=\mathbb{P}\!\left(\hat{y}=y \mid x,\pi\right),
\label{eq:cap_as_prob}
\end{equation}
where the probability is induced by the model's generation distribution and the parsing rule.
This emphasizes a key point for deployment: ``capability'' is not a protocol-free constant, but a quantity conditional on the interaction setting.

In real systems, protocols do not remain fixed.
A more deployment-faithful target is an expectation over both the task distribution $\mathcal{P}(x,y)$ and a distribution of \emph{reasonable} protocols $\mathcal{P}(\pi)$:
\begin{equation}
\mathrm{Cap}(m)
=
\mathbb{E}_{(x,y)\sim\mathcal{P}}
\ \mathbb{E}_{\pi\sim\mathcal{P}(\pi)}
\big[c_m(x,y;\pi)\big].
\label{eq:cap_target}
\end{equation}
Equivalently, using Eq.~\eqref{eq:cap_as_prob},
\begin{equation}
\mathrm{Cap}(m)
=
\mathbb{E}_{\pi\sim\mathcal{P}(\pi)}
\ \mathbb{E}_{(x,y)\sim\mathcal{P}}
\Big[\mathbb{P}\!\left(\hat{y}=y \mid x,\pi\right)\Big].
\label{eq:cap_target_prob}
\end{equation}
A single-score leaderboard under one fixed protocol $\pi_0$ corresponds to the narrow slice
$\mathbb{E}_{(x,y)\sim\mathcal{P}}\!\left[c_m(x,y;\pi_0)\right]$,
which ignores protocol uncertainty and drift. As a result, it can conflate models that are strong under $\pi_0$ but highly protocol-sensitive with models that are broadly robust.

To formalize ``sensitivity,'' we can treat protocols as having a continuous representation.
Let $\phi(\pi)\in\mathbb{R}^d$ be a feature vector of protocol attributes (e.g., strength of format constraints, degree of caution, conciseness requirements), and write
$c_m(x,y;\pi)=\tilde{c}_m\!\left(x,y;\phi(\pi)\right)$.
Then a natural notion of local protocol sensitivity is the gradient
\begin{equation}
\nabla_{\phi}\, \tilde{c}_m\!\left(x,y;\phi(\pi)\right).
\label{eq:grad_phi}
\end{equation}
Along a one-dimensional perturbation path $\pi(\theta)$ (with perturbation intensity $\theta$), local slope can be expressed by a partial derivative:
\begin{equation}
s_m(x,y;\theta)
=
\frac{\partial}{\partial \theta}\,
\tilde{c}_m\!\left(x,y;\phi(\pi(\theta))\right).
\label{eq:local_sensitivity}
\end{equation}
Here $s_m(x,y;0)$ captures the instantaneous effect of small protocol shifts.

\section{Method}

\subsection{CreditAudit Design}
CreditAudit turns protocol variation from an implicit nuisance factor into a controlled, aligned, and comparable audit dimension.
We evaluate a set of models $\mathcal{M}$ across a set of system-prompt templates (scenarios) $\mathcal{T}=\{1,\dots,T\}$ and benchmarks $\mathcal{B}=\{1,\dots,K\}$.
The template set is designed to satisfy two constraints.
First, semantic alignment: the same template index $t$ expresses the same protocol intent across benchmarks (e.g., ``output only the option letter,'' ``be cautious and verify,'' ``be concise,'' ``follow a strict format''), enabling cross-task comparability.
Second, non-adversariality: templates represent benign, routine protocol variations rather than sabotage-style instructions.

For each benchmark $b\in\mathcal{B}$, we define a fixed evaluation subset
\begin{equation}
\mathcal{D}_b=\{(x_{b,i}, \mathcal{C}_{b,i}, y_{b,i})\}_{i=1}^{N_b},
\label{eq:dataset}
\end{equation}
where $x_{b,i}$ is the question stem, $\mathcal{C}_{b,i}$ is the choice set, and $y_{b,i}\in\mathcal{C}_{b,i}$ is the gold answer.
Using fixed subsets isolates protocol-induced variation from sampling noise, so that the measured fluctuations primarily reflect changes in $\pi$ rather than changes in the evaluated items.
For each tuple $(m,t,b,i)$, the model generates
\begin{equation}
r_{m,t,b,i}\sim p_m(\cdot\mid x_{b,i},\pi_t).
\label{eq:gen}
\end{equation}
This yields a complete model $\times$ template $\times$ benchmark evaluation cube, which serves as the audit evidence for robustness estimation.

\subsection{Parsing and Benchmark Scores: Auditability and Reproducibility}
Even for multiple-choice benchmarks, model outputs are typically free text.
We define a parsing function $g(\cdot)$ that maps a response to a discrete option:
\begin{equation}
\hat{y}_{m,t,b,i}=g(r_{m,t,b,i}),
\qquad
\hat{y}_{m,t,b,i}\in\mathcal{C}_{b,i}.
\label{eq:parse}
\end{equation}
Per-item correctness is then
\begin{equation}
h(r_{m,t,b,i}, y_{b,i})
=
\mathbb{1}\!\left(\hat{y}_{m,t,b,i}=y_{b,i}\right),
\label{eq:item_acc}
\end{equation}
and the benchmark score is empirical accuracy:
\begin{equation}
S_{m,t,b}
=
\frac{1}{N_b}\sum_{i=1}^{N_b}
\mathbb{1}\!\left(\hat{y}_{m,t,b,i}=y_{b,i}\right).
\label{eq:bench_score}
\end{equation}
We choose ``parsing + accuracy'' rather than an additional judge model to preserve auditability: the scoring rule is transparent and verifiable, and fluctuations cannot be attributed to evaluator instability.
Under this setup, $S_{m,t,b}$ can be interpreted as an empirical estimate of the protocol-conditional correctness probability averaged over the benchmark subset.

\subsection{From One Leaderboard Score to Two Dimensions: Mean Ability and Stability Risk}
For each model $m$ under template $t$, we aggregate $K$ benchmark scores with equal weights to form an overall score:
\begin{equation}
S_{m,t}
=
\frac{1}{K}\sum_{b=1}^{K} S_{m,t,b}.
\label{eq:overall}
\end{equation}
The sequence $\{S_{m,t}\}_{t=1}^{T}$ defines the model's trajectory under standard system-prompt variations.
We define two model-level statistics:
\begin{align}
\mu_m
&=
\frac{1}{T}\sum_{t=1}^{T} S_{m,t},
\label{eq:mu}
\\
\sigma_m
&=
\sqrt{
\frac{1}{T-1}\sum_{t=1}^{T}\left(S_{m,t}-\mu_m\right)^2
}.
\label{eq:sigma}
\end{align}
$\mu_m$ captures average performance across scenarios, while $\sigma_m$ captures sensitivity to benign protocol variation, which we interpret as a stability-risk signal.

To connect $\sigma_m$ to protocol-space sensitivity, consider $S_m(\pi)$ as a random variable under $\Pi\sim\mathcal{P}(\pi)$, observed at discrete samples $\{\pi_t\}$.
Under a continuous representation $\phi(\pi)$ and local differentiability, a first-order approximation around a reference protocol $\pi_0$ gives
\begin{equation}
S_m(\pi)\approx S_m(\pi_0)+\nabla_{\phi}S_m(\pi_0)^\top(\phi(\pi)-\phi(\pi_0)).
\label{eq:taylor}
\end{equation}
This yields the standard linear variance-propagation approximation:
\begin{equation}
\mathrm{Var}_{\pi\sim\mathcal{P}(\pi)}\!\big[S_m(\pi)\big]
\ \approx\
\nabla_{\phi}S_m(\pi_0)^\top\,
\mathrm{Cov}\!\big(\phi(\Pi)\big)\,
\nabla_{\phi}S_m(\pi_0),
\label{eq:var_grad}
\end{equation}
providing an interpretation: larger gradients in protocol space imply larger score fluctuations under reasonable protocol perturbations, all else equal. CreditAudit estimates this variability empirically via $\sigma_m$ over an aligned set of benign templates.

For diagnostic purposes, benchmark-specific mean and volatility can also be computed:
\begin{equation}
\mu_{m,b}=\frac{1}{T}\sum_{t=1}^{T}S_{m,t,b},
\qquad
\sigma_{m,b}=
\sqrt{\frac{1}{T-1}\sum_{t=1}^{T}(S_{m,t,b}-\mu_{m,b})^2 }.
\label{eq:bench_mu_sigma}
\end{equation}
These decompositions help identify whether overall instability is broad-based or driven by a particular benchmark type that may align more closely with the target workload.

\subsection{Credit Grades as Interpretable ``Risk Labels''}
To provide an interpretable stability label, CreditAudit discretizes $\sigma_m$ by cross-model quantiles.
Let $q_{0.25},q_{0.50},q_{0.75}$ be the 25th, 50th, and 75th percentiles of $\{\sigma_m\}_{m\in\mathcal{M}}$.
Define the credit grade:
\begin{equation}
G(m)=
\begin{cases}
\mathrm{AAA}, & \sigma_m \le q_{0.25},\\
\mathrm{AA},  & q_{0.25}<\sigma_m \le q_{0.50},\\
\mathrm{A},   & q_{0.50}<\sigma_m \le q_{0.75},\\
\mathrm{BBB}, & \sigma_m > q_{0.75}.
\end{cases}
\label{eq:grade}
\end{equation}
Higher grades indicate greater stability under prompt variation.
In the main results, the grade is reported together with $(\mu_m,\sigma_m)$ to support downstream model selection, tiered deployment, and disciplined allocation of testing effort across deployment regimes.

\section{Experimental}
\label{sec:exp}

\subsection{Setup}

\paragraph{\textbf{Benchmarks and Task Format}}
We evaluate on three public multiple-choice benchmarks: GPQA~\cite{rein2023gpqa}, TruthfulQA\cite{lin2022truthfulqa}, and MMLU-Pro\cite{wang2024mmlupro}.
These benchmarks provide complementary coverage of high-difficulty reasoning, truthfulness judgments, and broad knowledge testing.
We focus on the multiple-choice format because it supports a transparent and verifiable scoring rule (accuracy), which is essential for attributing performance variation to prompt-induced effects rather than to subjective grading.

\paragraph{\textbf{Sampling and Shared Evaluation Subsets}}
Running a full factorial evaluation across many models and many prompt templates can be expensive.
For each benchmark $b$, we construct a fixed evaluation subset $\mathcal{D}_b$ of size $N_b$ (Eq.~\eqref{eq:dataset}) using a deterministic sampling procedure driven by a fixed random seed.
Within an experimental run, all models and all prompt templates are evaluated on the same $\mathcal{D}_b$, ensuring strict horizontal comparability.
This design is particularly important for estimating stability risk $\sigma_m$, because it isolates protocol-induced variation from sample-induced variation.

\paragraph{\textbf{System Prompt Templates as Controlled Protocol Perturbations}}
We construct a set of $T$ standard system-prompt templates to represent normal variations in real deployments (e.g., enforcing ``output only the option letter'', emphasizing conciseness, requiring cautious verification, etc.).
The templates are designed as usable, non-adversarial system instructions; they do not include sabotage-like content intended to reduce accuracy.
For each template index $t$, we provide benchmark-specific system prompts aligned in style, so that the same template index corresponds to comparable protocol intent across all benchmarks.
For each model, the resulting scores $\{S_{m,t}\}_{t=1}^{T}$ define the prompt-induced performance trajectory used to compute $(\mu_m,\sigma_m)$.

\paragraph{\textbf{Evaluation Procedure and Aggregation}}
For every model $m$ and prompt template $t$, we run MC evaluation on each benchmark $b$ and compute $S_{m,t,b}$ as accuracy (Eq.~\eqref{eq:bench_score}).
We then compute the overall score $S_{m,t}$ by equal-weight averaging across benchmarks (Eq.~\eqref{eq:overall}), followed by model-level mean ability and stability risk (Eqs.~\eqref{eq:mu}--\eqref{eq:sigma}).
Finally, we assign a credit grade $G(m)$ based on the cross-model quantiles of $\sigma_m$ (Eq.~\eqref{eq:grade}).
In addition to overall statistics, we compute benchmark-specific $(\mu_{m,b},\sigma_{m,b})$ (Eq.~\eqref{eq:bench_mu_sigma}) for diagnostic analysis.

\subsection{Average performance and scenario robustness jointly characterize model quality}
Table~\ref{tab:main} reports CreditAudit outcomes for all evaluated models.
Each row contains (i) the overall average score $\mu_m$ and scenario-induced score fluctuation $\sigma_m$ across a family of semantically aligned system-instruction scenarios (Eqs.~\eqref{eq:mu}--\eqref{eq:sigma}), and (ii) benchmark-specific average/fluctuation pairs $(\mu_{m,b},\sigma_{m,b})$ (Eq.~\eqref{eq:bench_mu_sigma}) for GPQA, TruthfulQA, and MMLU-Pro.
Unlike conventional single-score leaderboards, this layout makes two deployment-relevant axes simultaneously visible: \emph{ability level} (how strong a model is on average) and \emph{scenario sensitivity} (how much the model moves under routine protocol variation).

Credit grades are assigned by cross-model quantiles of overall fluctuation $\sigma_m$:
\[
q_{0.25}=1.30,\quad q_{0.50}=1.57,\quad q_{0.75}=2.04,
\]
so AAA denotes the most robust quartile and BBB the least robust quartile (Eq.~\eqref{eq:grade}).
To support selection-oriented reading, Table~\ref{tab:main} is sorted by overall fluctuation $\sigma_m$ from low to high (most robust at the top).

Two observations from Table~\ref{tab:main} are especially important for deployment.
First, \textbf{similar mean scores do not imply similar robustness}.
Models that appear comparable under a single-score view can differ markedly in $\sigma_m$, meaning their operational behavior may diverge once prompts, output constraints, or interaction modes shift in normal (non-adversarial) ways.
Second, \textbf{robustness can be task-dependent}.
The benchmark-level decompositions $(\mu_{m,b},\sigma_{m,b})$ reveal whether a model's overall volatility is broad-based or driven disproportionately by a specific benchmark type.
This matters because real applications rarely match a uniform mixture of tasks: a model that is stable overall but fragile on a benchmark aligned with the target workload can still be risky in practice.

\begin{table*}[t]
\centering
\caption{Main CreditAudit results, sorted by scenario-induced score fluctuation.}
\label{tab:main}
\small
\setlength{\tabcolsep}{4pt} 
\renewcommand{\arraystretch}{1.05}

\resizebox{\textwidth}{!}{%
\begin{tabular}{l c cc cc cc cc}
\toprule
Model & Grade &
Overall avg $\mu$ & Fluct.\ $\sigma$ &
GPQA avg $\mu$ & Fluct.\ $\sigma$ &
TruthfulQA avg $\mu$ & Fluct.\ $\sigma$ &
MMLU-Pro avg $\mu$ & Fluct.\ $\sigma$ \\
\midrule
\texttt{bytedance-seed/seed-1.6-flash} & AAA & 70.77 & 0.63 & 64.9 & 1.91 & 70.4 & 2.72 & 77.0 & 1.49 \\
\texttt{google/gemini-2.5-pro} & AAA & 80.13 & 0.72 & 66.7 & 2.67 & 93.4 & 1.51 & 80.3 & 1.25 \\
\texttt{bytedance-seed/seed-1.6} & AAA & 81.87 & 1.24 & 76.2 & 2.53 & 88.3 & 1.42 & 81.1 & 1.29 \\
\texttt{qwen/qwen3-32b} & AAA & 59.13 & 1.30 & 37.2 & 2.15 & 78.0 & 3.40 & 62.2 & 2.90 \\
\texttt{qwen/qwen3-235b-a22b-2507} & AA & 62.30 & 1.43 & 50.3 & 1.95 & 79.9 & 1.85 & 56.7 & 2.71 \\
\texttt{z-ai/glm-4.5} & AA & 66.80 & 1.51 & 47.2 & 3.77 & 84.1 & 3.45 & 69.1 & 3.84 \\
\texttt{moonshotai/kimi-k2-0905} & AA & 63.97 & 1.57 & 48.9 & 2.33 & 83.5 & 1.90 & 59.5 & 3.06 \\
\texttt{deepseek/deepseek-chat-v3-0324} & A & 58.53 & 1.79 & 46.8 & 4.39 & 72.3 & 3.53 & 56.5 & 1.51 \\
\texttt{deepseek/deepseek-v3.2} & A & 57.13 & 1.81 & 42.3 & 3.59 & 72.4 & 1.58 & 56.7 & 3.47 \\
\texttt{meta-llama/llama-3.3-70b-instruct} & A & 52.20 & 2.04 & 40.7 & 3.27 & 72.4 & 2.72 & 43.5 & 3.17 \\
\texttt{meta-llama/llama-3-8b-instruct} & BBB & 30.17 & 2.09 & 27.4 & 4.03 & 38.3 & 3.33 & 24.8 & 4.92 \\
\texttt{z-ai/glm-4.5-air} & BBB & 54.80 & 2.25 & 40.7 & 3.74 & 77.9 & 2.69 & 45.8 & 1.75 \\
\texttt{google/gemini-2.5-flash-lite} & BBB & 67.27 & 2.63 & 55.1 & 5.78 & 78.1 & 2.81 & 68.6 & 11.07 \\
\bottomrule
\bottomrule
\end{tabular}%
}
\end{table*}

\subsection{Model selection is a risk problem}
CreditAudit is designed for \emph{model selection}, not for leaderboard ranking.
In many deployments, the main cost is not ``a few points of average accuracy'' but \textbf{unreliable behavior when the interaction scenario changes in normal ways}.
This is why our core output is a credit-style robustness label (Eq.~\eqref{eq:grade}), and why \textbf{scenario-induced score fluctuation} $\sigma_m$ is treated as first-class evidence rather than a side statistic.

This perspective matches how real systems accumulate cost.
When a model is sensitive to routine protocol changes, teams pay for that sensitivity through additional prompt hardening, increased regression-testing frequency, larger scenario coverage requirements, more conservative rollouts, and heavier monitoring and human fallback.
These costs are often dominated by \emph{rare-but-severe} failures (format violations that break parsers, inconsistent refusals that disrupt flows, wrong tool routing that derails a chain), rather than by small differences in average benchmark accuracy.

\paragraph{Real systems constantly rotate through scenarios.}
Deployed systems rarely keep one fixed instruction context.
An application may shift scenarios when it moves between tools, changes output schema, enters a verification mode, or adapts to user preferences.
These are normal operational adjustments, not adversarial attacks.
If a model reacts strongly to such scenario shifts, it can become unpredictable even when the underlying user intent is unchanged.

\paragraph{Agentic workflows amplify scenario effects.}
For agent settings, the model is invoked repeatedly and the pipeline contains many decision points (routing, tool calls, schema compliance, self-checking).
Here, robustness can matter more than peak score:
a slightly weaker but \emph{steady} model can outperform a higher-score but \emph{scenario-fragile} model at the system level, because occasional deviations break the trajectory (wrong tool choice, schema violations, refusal oscillations, etc.).
Accordingly, CreditAudit encourages a selection rule closer to: \textbf{first pick low fluctuation, then optimize score within that robustness tier.}
This rule is not meant to be universal, but it captures a common deployment regime: as the pipeline becomes longer and more dynamic, instability becomes the dominant driver of end-to-end cost.

\paragraph{Scores are evidence; grades are decisions.}
A score summarizes expected performance under an evaluation slice.
A credit grade summarizes the \emph{risk} of relying on the model when scenarios vary in normal ways.
This is why Table~\ref{tab:main} is sorted by fluctuation and why our main visualization uses a score--fluctuation plane rather than score alone.
In other words, $\mu$ supports capability comparison, while $\sigma$ and the corresponding grade support risk-aware prioritization.

\subsection{Score-fluctuation map: the selection frontier}
\begin{figure}[t]
\centering
\includegraphics[width=\linewidth]{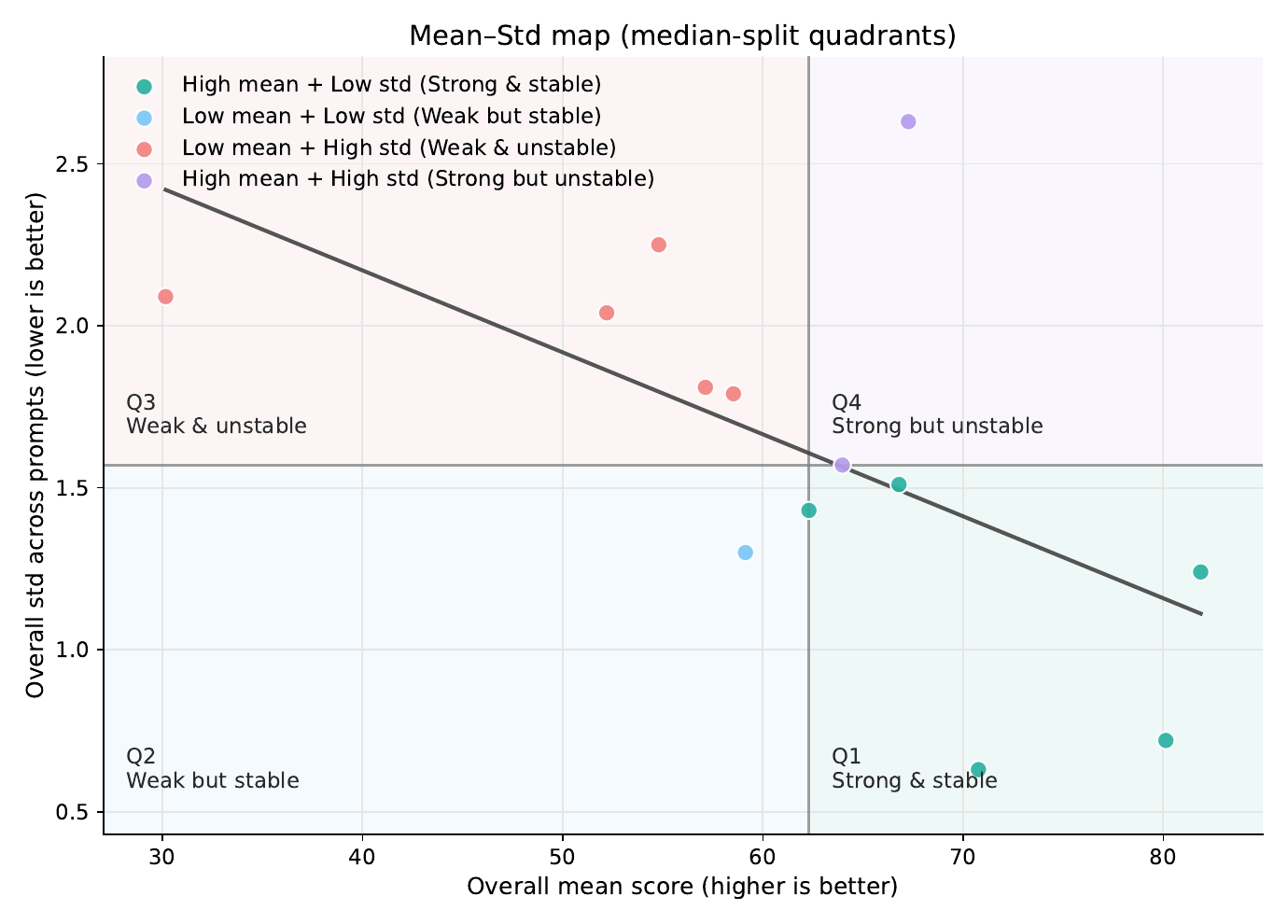}
\caption{Overall score vs.\ scenario-induced fluctuation with median-split quadrants.}
\label{fig:overall_map}
\end{figure}

Figure~\ref{fig:overall_map} places each model in a two-dimensional space: overall average score (higher is better) and overall scenario-induced score fluctuation (lower is better).
We split the plane by the medians (vertical: median score $=62.3$; horizontal: median fluctuation $=1.57$), yielding four quadrants that translate into concrete selection regimes.
This visualization makes explicit a practical frontier: models compete not only on average performance, but also on how predictably they behave when the protocol shifts.

\paragraph{Q1: High score \& low fluctuation: ``safe default''.}
Models in the bottom-right combine strong average performance with robustness to normal scenario changes.
They are the most defensible defaults for production and agentic settings.
In our results, Seed-1.6, Gemini-2.5-Pro, and Seed-1.6-Flash are clear Q1 models (all AAA).
Operationally, Q1 models reduce integration friction because engineers can iterate on wording and constraints with less fear of triggering large quality swings.

\paragraph{Q4: High score but high fluctuation: ``strong but scenario-fragile''.}
This quadrant contains models that look good on average but move a lot when scenarios shift.
Gemini-2.5-Flash-Lite is the clearest example (BBB): it is above-median in score yet the most volatile overall.
For tightly controlled single-shot usage this may be acceptable; for agents or multi-step workflows it signals higher operational risk.
In practice, Q4 models typically require heavier scenario testing and stricter protocol control to keep failure rates within acceptable bounds.

\paragraph{Q2: Lower score but low fluctuation: ``predictable baseline''.}
Q2 models are not top performers on average but are comparatively robust.
This quadrant matters because many real systems care more about predictability than peak score.
For instance, Qwen3-32B receives AAA due to robustness even though its score is below the median.
Such models can be attractive in budget-constrained regimes where stability supports simpler monitoring and a more consistent user experience.

\paragraph{Q3: Lower score and high fluctuation: ``avoid by default''.}
Models here combine weaker average performance with instability, making them difficult to rely on when the application must tolerate scenario shifts.
They tend to be costly to integrate because teams pay both in baseline performance and in risk mitigation effort.

\paragraph{Selection rule of thumb.}
If the application is agentic or high-stakes, prioritize \textbf{low fluctuation first}, then consider score within the acceptable grade tier.
If the application is single-shot and tightly scenario-controlled, score can weigh more heavily, but Q4 still requires caution.
This is precisely the kind of regime-dependent trade-off that single-score leaderboards cannot communicate.

\subsection{Measured perturbation under scenario shifts: model $\times$ scenario structure}
\begin{figure}[t]
\centering
\includegraphics[width=\linewidth]{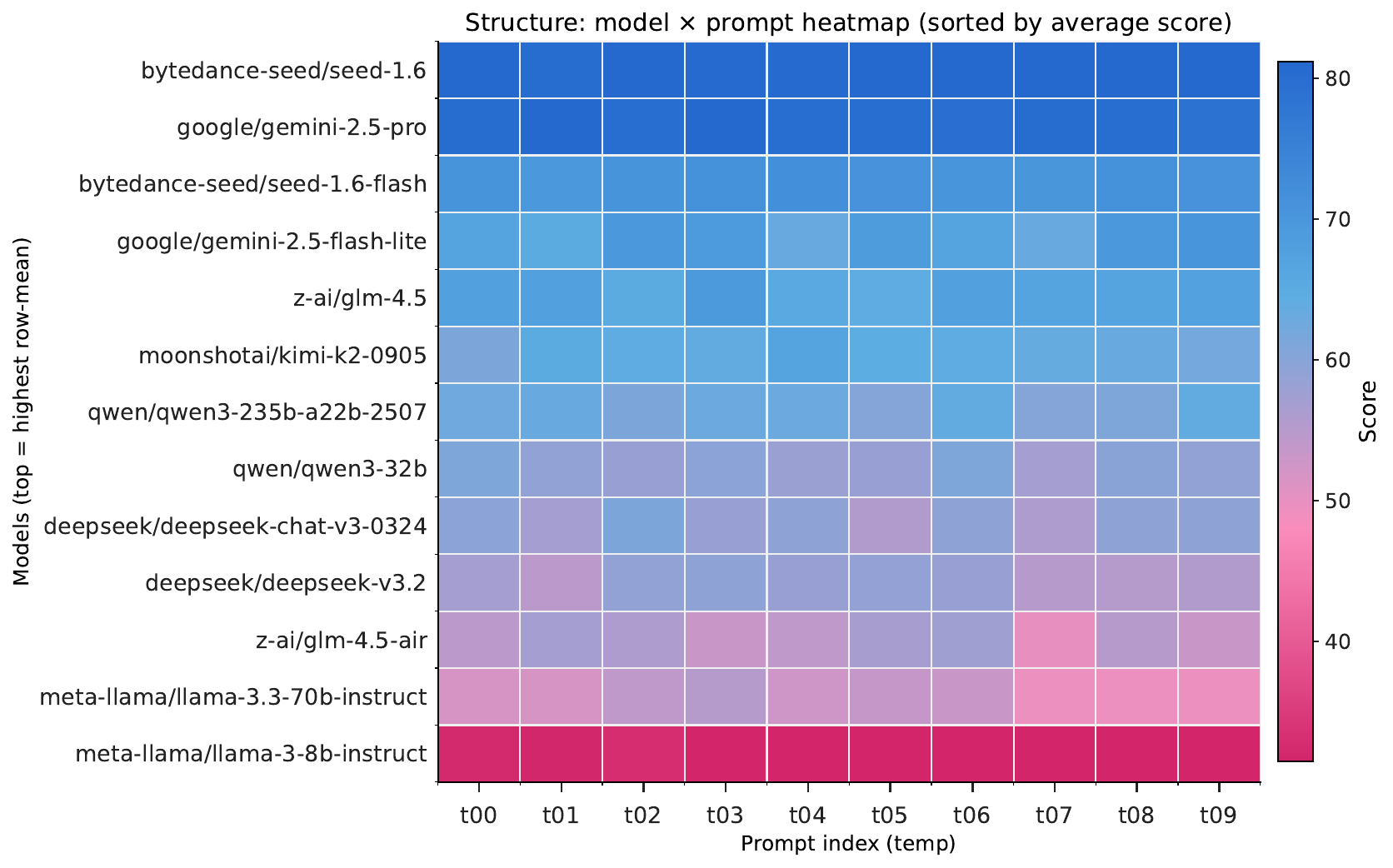}
\caption{Model $\times$ scenario heatmap of overall scores $S_{m,t}$.}
\label{fig:heatmap_structure}
\end{figure}

Figure~\ref{fig:heatmap_structure} visualizes the full model $\times$ scenario matrix of overall scores $S_{m,t}$ (Eq.~\eqref{eq:overall}).
For a selection-oriented audit, this is direct evidence of how much a model moves when the system-instruction scenario changes.
Beyond summarizing movement with $\sigma_m$, the matrix reveals \emph{where} instability concentrates: whether a model is uniformly sensitive across scenarios, or whether it is mostly stable but collapses under a small subset of protocol variants.
This distinction matters for deployment because scenario-specific collapses correspond to tail risk, which is often amplified in multi-step pipelines.

\paragraph{No "easy scenario" effect; the signal is model-specific fluctuation.}
A potential concern is whether certain scenarios consistently increase or decrease scores, making fluctuations more challenging to interpret.
However, our neutrality check (Appendix~\ref{sec:appendix_prompt_neutral}) indicates that average scores across models are nearly uniform across scenarios.
Thus, the primary signal in Figure~\ref{fig:heatmap_structure} is not that "some scenarios are universally higher or lower," but rather the \textbf{degree of fluctuation} observed among different models under a set of semantically consistent scenario transformations.
This supports interpreting $\sigma_m$ as model-specific sensitivity rather than an artifact of globally drifting template difficulty.

\paragraph{Robustness as integration cost.}
A row with small within-row spread implies that a model is robust across reasonable scenario changes.
This reduces integration cost: engineers can iterate on system wording or enforce different schemas without triggering large shifts.
Conversely, larger within-row spread implies higher operational risk and higher testing burden, especially for agent pipelines.
From a deployment perspective, Figure~\ref{fig:heatmap_structure} therefore complements Table~\ref{tab:main}: the table supports prioritization and comparison, while the matrix supports diagnosis and targeted mitigation.

\paragraph{Credit interpretation.}
In credit terms, a model with smaller scenario-induced fluctuation is less likely to surprise you under routine regime changes.
This is why we sort Table~\ref{tab:main} by fluctuation and assign grades based on fluctuation quantiles.
The heatmap provides the underlying ``audit trail'' behind that grade: it is the concrete, scenario-by-scenario evidence that the model remains steady (or not) when the protocol shifts.

\subsection{A brief diagnostic note}
We include two supporting diagnostics in the appendix:
(i) a scenario neutrality check that verifies scenario variants do not introduce a systematic shift across models (Appendix~\ref{sec:appendix_prompt_neutral}),
and (ii) per-model score distributions across scenarios, which provides a distributional view of robustness (Appendix~\ref{sec:appendix_dist}).
Benchmark-level decompositions are also provided for completeness (Appendix~\ref{sec:appendix_bench_diag}).
Together, these diagnostics are intended to make the main results easier to trust and easier to act on: neutrality supports attributing $\sigma_m$ to model sensitivity rather than template drift, while distributions and benchmark-level views help practitioners identify tail-risk patterns and workload-specific fragility.

\section{Conclusion}
\label{sec:conclusion}

This paper argues that \emph{model selection is not only a score problem, but also a risk problem}.
In real deployments, a model is rarely used under a single fixed instruction context: system prompts evolve with product iteration, tool-use constraints, output schemas, and safety or verification modes.
As a result, two models with similar average benchmark scores can have very different \emph{operational reliability} once the interaction scenario changes in normal, non-adversarial ways.

To make this reliability visible and actionable, we introduced \textbf{CreditAudit}, a two-dimensional audit for priority LLM selection.
Instead of reporting one leaderboard number, CreditAudit evaluates each model across a family of semantically aligned system-prompt templates and reports
(i) \textbf{mean ability} $\mu_m$ as the average performance across scenarios, and
(ii) \textbf{scenario-induced score fluctuation} $\sigma_m$ as a direct estimate of stability risk under routine protocol variation.
We further translate $\sigma_m$ into an interpretable \emph{credit grade} (AAA--BBB) via cross-model quantiles, providing a decision-oriented ``risk label'' that complements capability scores.

Our empirical results on three public multiple-choice benchmarks (GPQA, TruthfulQA, and MMLU-Pro) highlight three takeaways.
First, \textbf{robustness varies substantially across models}: some models remain stable across scenario shifts while others exhibit large swings, even when average scores appear competitive.
Second, \textbf{the selection frontier is inherently two-dimensional}:
a ``safe default'' model is one that combines high $\mu_m$ with low $\sigma_m$, whereas a high-score but high-$\sigma_m$ model can be scenario-fragile and thus costly to integrate for agentic or multi-step workflows.
Third, our prompt-set neutrality diagnostic indicates that the scenarios do not act as globally ``easy'' or ``hard'' regimes across models; instead, the dominant signal is \emph{model-specific} sensitivity, supporting the interpretation of $\sigma_m$ as stability risk rather than an artifact of template difficulty.

Beyond the specific numbers reported in this paper, CreditAudit contributes a \textbf{reproducible evaluation pattern}:
(1) treat the interaction protocol as part of the evaluation environment,
(2) measure both performance and sensitivity under controlled, benign protocol perturbations, and
(3) output an interpretable robustness label to support selection decisions.
We will release the system prompt templates and evaluation toolkit to facilitate extensions to additional benchmarks, broader protocol families (e.g., tool-use and schema constraints), and alternative aggregation rules.

\paragraph{Limitations and future work.}
CreditAudit currently focuses on multiple-choice benchmarks to preserve transparent and auditable scoring.
Future work should extend the audit to open-ended generation tasks with carefully designed, verifiable rubrics, and to application-level metrics (e.g., tool-call correctness, schema compliance, and multi-step success rate) where scenario sensitivity may compound.
In addition, while we use a fixed set of prompt templates as a practical approximation of ``reasonable'' scenario variation, broader and domain-specific scenario families could further improve coverage.
Finally, robustness is only one axis of deployment readiness; cost, latency, safety behavior, and refusal patterns should be integrated into a multi-objective selection framework.

Overall, CreditAudit reframes LLM selection as choosing not only \emph{how capable} a model is on average, but also \emph{how dependable} it remains when the system inevitably changes around it.


\bibliographystyle{ACM-Reference-Format}
\bibliography{sample}


\appendix

\section{Additional Diagnostics}
\label{sec:appendix}

\subsection{Scenario neutrality check}
\label{sec:appendix_prompt_neutral}

CreditAudit interprets scenario-induced score fluctuation $\sigma_m$ as \emph{model-specific sensitivity} to routine system-prompt variations.
This interpretation requires that the scenario set itself is broadly ``neutral'': i.e., templates should not systematically make the evaluation globally easier or harder across all models.
Otherwise, a large $\sigma_m$ could partly reflect an artifact of scenario difficulty drift rather than a model's robustness.

\paragraph{Definition.}
For each scenario (template) index $t$, we compute the across-model mean score
\begin{equation}
\bar{S}_t=\frac{1}{|\mathcal{M}|}\sum_{m\in\mathcal{M}}S_{m,t},
\label{eq:appendix_prompt_effect}
\end{equation}
where $S_{m,t}$ is the overall score of model $m$ under scenario $t$ (Eq.~\eqref{eq:overall}).
If the template set is neutral, $\{\bar{S}_t\}$ should be approximately flat across $t$, up to small sampling noise.

\begin{figure}[!htbp]
\centering
\includegraphics[width=1.0\linewidth]{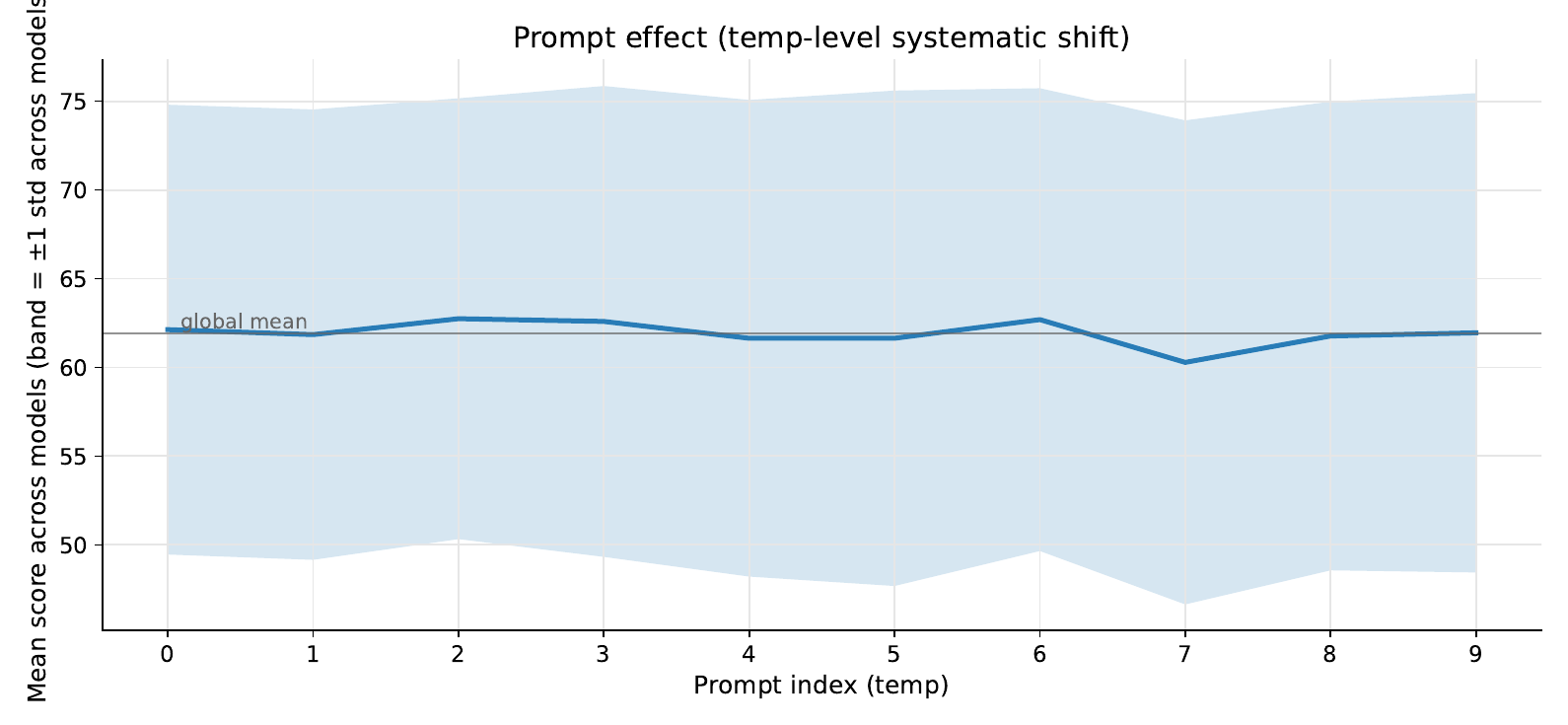}
\caption{Across-model mean score $\bar{S}_t$ by scenario (template). A near-flat trend indicates that scenario variants do not introduce a strong systematic shift in overall difficulty across models.}
\label{fig:prompt_effect}
\end{figure}

\paragraph{Reading the diagnostic.}
Figure~\ref{fig:prompt_effect} plots $\bar{S}_t$ across all scenarios.
The near-flat trend indicates that scenario variants do not induce a strong systematic shift in overall difficulty.
This supports our main claim in Section~\ref{sec:main} that the primary signal captured by $\sigma_m$ is \emph{row-wise fluctuation} (model-specific movement across scenarios), rather than a column-wise ``easy scenario'' or ``hard scenario'' effect.

\paragraph{Why this matters for selection.}
In the main table (Table~\ref{tab:main}), models are sorted by $\sigma_m$ and assigned credit grades based on cross-model quantiles of $\sigma_m$ (Eq.~\eqref{eq:grade}).
The neutrality check justifies treating those grades as a stability-risk label:
if scenarios were globally drifting in difficulty, a low grade could partly be an artifact of template design.
By contrast, with neutrality, a low grade more credibly reflects that the model is sensitive to normal instruction changes, which increases integration cost and operational risk in real systems.

\paragraph{Failure modes and practical guidance.}
This diagnostic can fail in two common ways:
(i) \textbf{global difficulty drift} where certain templates systematically boost or depress most models (a strong slope or step in $\bar{S}_t$), and
(ii) \textbf{interaction mismatch} where a subset of templates changes the task (e.g., injects extra constraints that effectively alter what is being evaluated).
If either occurs, we recommend revising the template set (removing or rewriting the problematic scenarios) and re-running the audit, because $\sigma_m$ would then mix ``model sensitivity'' with ``scenario difficulty shift''.
In our experiments, the observed near-flat $\bar{S}_t$ suggests that the scenario set functions as intended: semantically aligned, benign protocol perturbations.

\subsection{Distribution evidence: per-model score distributions across scenarios}
\label{sec:appendix_dist}

\begin{figure}[!htbp]
\centering
\includegraphics[width=\linewidth]{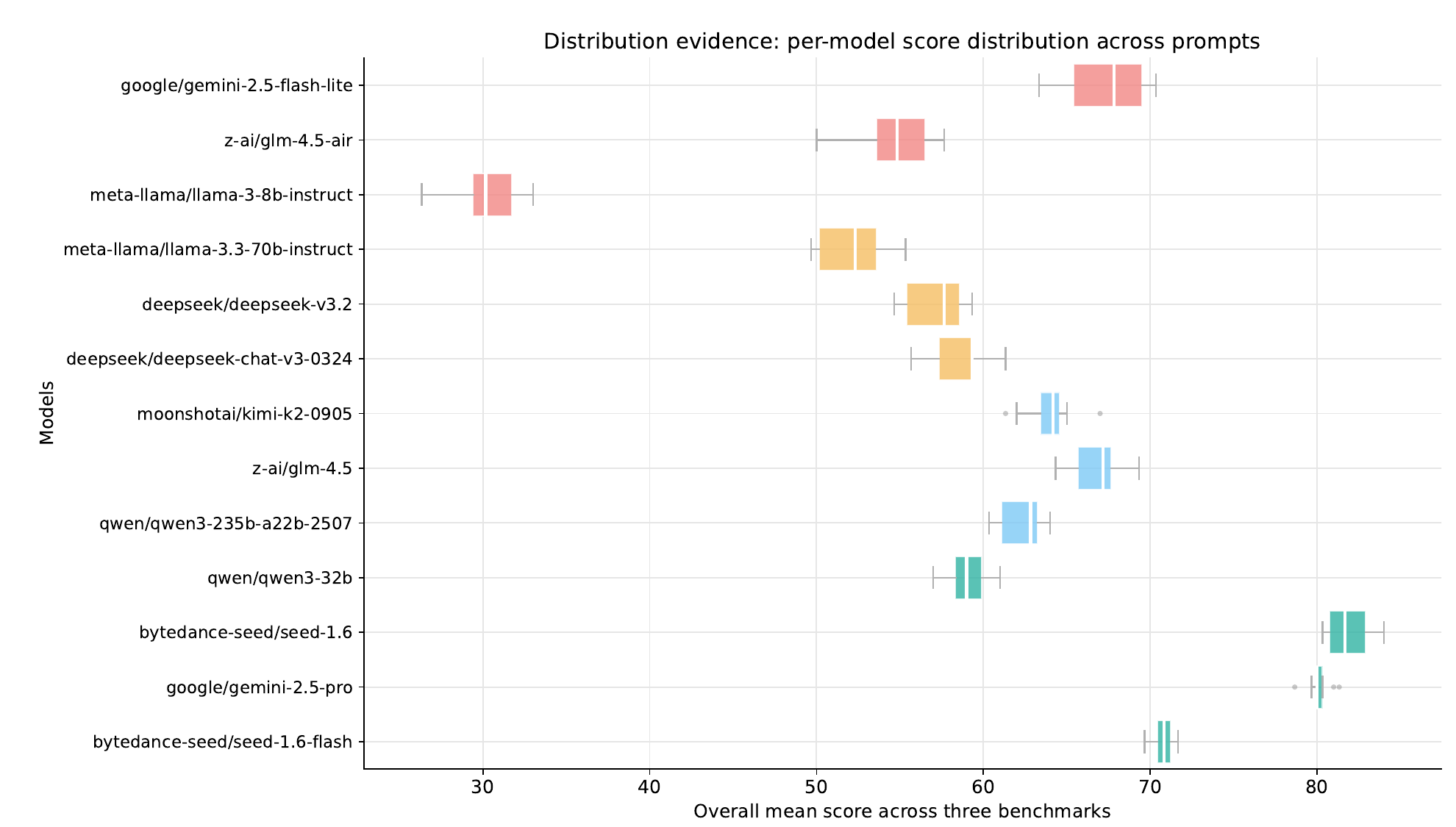}
\caption{Per-model distributions of overall scores across scenarios. Boxplots reveal not only average and variance but also tail behavior and scenario-specific collapses, complementing summary statistics $(\mu_m,\sigma_m)$.}
\label{fig:dist_box}
\end{figure}

While the main text summarizes robustness using $(\mu_m,\sigma_m)$, distributional evidence helps distinguish \emph{how} instability arises.
Two models can have similar $\sigma_m$ but very different distribution shapes: one may be consistently moderate (tight distribution), while another may be mostly strong but occasionally collapses under specific scenarios (heavy tail).
Such differences matter for deployment, especially for agentic workflows where rare failures can break multi-step trajectories.

\paragraph{What the plot shows.}
Figure~\ref{fig:dist_box} visualizes the distribution of overall scores $\{S_{m,t}\}_{t=1}^{T}$ for each model (boxplot).
The median and interquartile range show typical scenario performance, while outliers indicate sensitivity to specific scenario variants.
This complements the mean--fluctuation view in Figure~\ref{fig:overall_map}:
\begin{itemize}
    \item Models with \textbf{small IQR and short whiskers} correspond to low $\sigma_m$ and behave steadily under routine prompt changes, implying lower integration cost.
    \item Models with \textbf{long whiskers or pronounced outliers} may look competitive on average but exhibit scenario-specific failures, which is consistent with the ``scenario-fragile'' quadrant in the main text.
\end{itemize}

\paragraph{Operational interpretation.}
For single-shot, tightly controlled deployments, occasional outliers may be manageable through prompt hardening or targeted tests.
For agentic or multi-step pipelines, tail risk is amplified: a single scenario-induced failure can propagate (e.g., schema violation leads to a downstream parsing error), making models with heavy tails more expensive to validate and maintain.
Thus, the distribution view provides additional justification for treating robustness as a first-class axis in selection, rather than a secondary statistic.

\begin{figure}[htbp]
\centering
\includegraphics[width=0.9\linewidth]{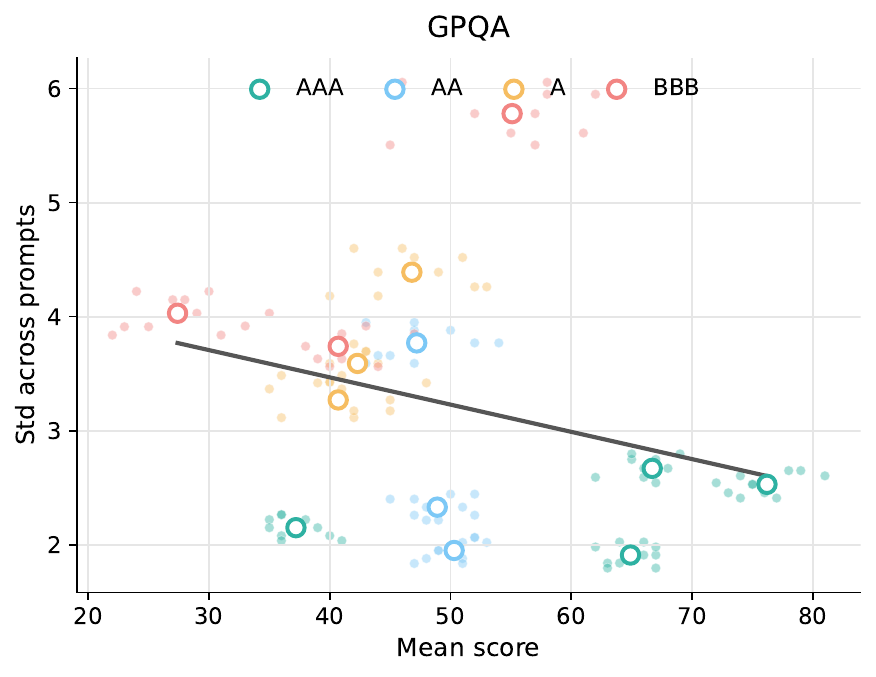}
\caption{GPQA score-fluctuation map using $(\mu_{m,\mathrm{GPQA}}, \sigma_{m,\mathrm{GPQA}})$. Center markers show per-model means and fluctuations; semi-transparent clouds show per-scenario realizations, highlighting scenario sensitivity on reasoning-intensive items.}
\label{fig:bench_map_gpqa}
\end{figure}

\paragraph{Why benchmark-level views are needed.}
A model's overall fluctuation $\sigma_m$ can be disproportionately driven by a single benchmark when that benchmark interacts more strongly with particular scenario styles, such as caution and verification prompts, strict output constraints, or changes in instruction verbosity.
As a result, aggregation can obscure where instability originates and which capability axis is most sensitive to scenario variation.
Benchmark-level views isolate these effects and answer practical questions:
\begin{itemize}
    \item Is a model robust across task types, or only robust on certain benchmarks?
    \item Does instability primarily arise on reasoning-intensive items (e.g., GPQA) or on truthfulness-oriented judgments (e.g., TruthfulQA)?
    \item Are there models that look stable overall but become scenario-fragile on a benchmark that is central to the target application domain?
\end{itemize}

\begin{figure}[htbp]
\centering
\includegraphics[width=0.9\linewidth]{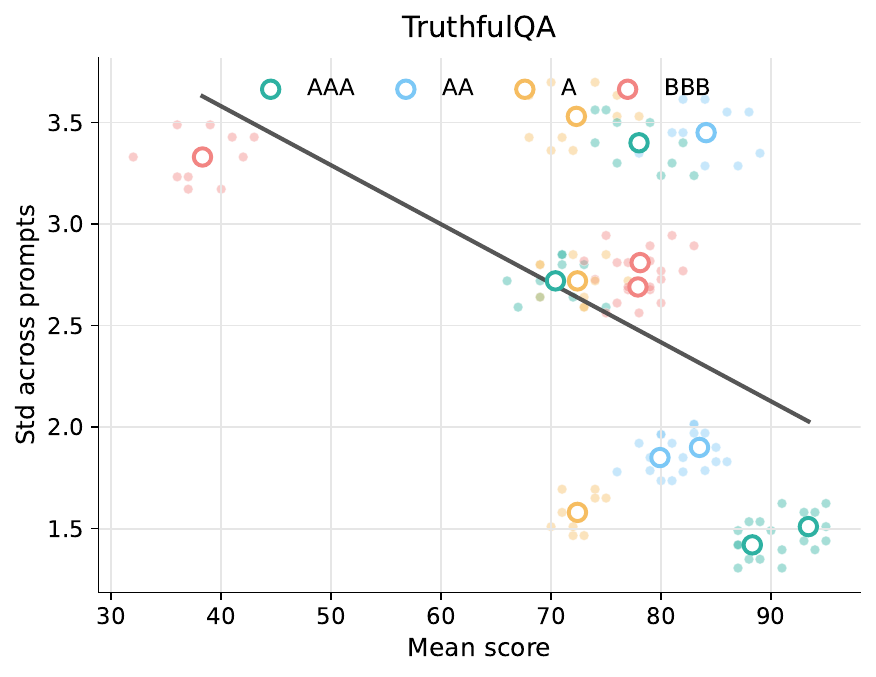}
\caption{TruthfulQA score-fluctuation map using $(\mu_{m,\mathrm{TruthfulQA}}, \sigma_{m,\mathrm{TruthfulQA}})$. Per-scenario clouds reveal whether robustness holds under style changes that affect truthfulness-oriented judgments.}
\label{fig:bench_map_truthfulqa}
\end{figure}

\begin{figure}[htbp]
\centering
\includegraphics[width=0.9\linewidth]{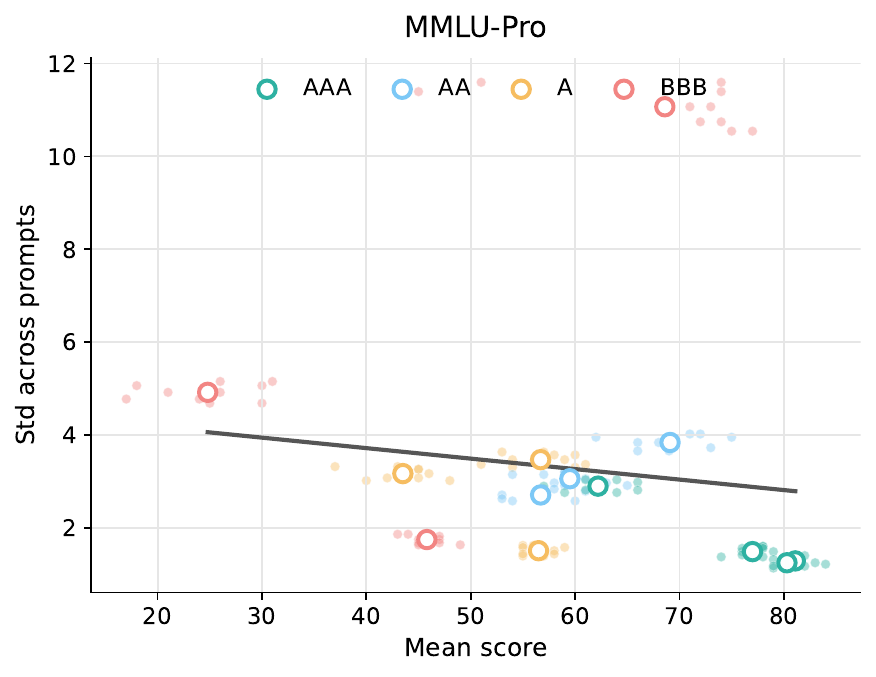}
\caption{MMLU-Pro score-fluctuation map using $(\mu_{m,\mathrm{MMLU\mbox{-}Pro}}, \sigma_{m,\mathrm{MMLU\mbox{-}Pro}})$. This benchmark-level view separates general knowledge stability from scenario-driven variance that may be masked by overall aggregation.}
\label{fig:bench_map_mmlu_pro}
\end{figure}

\paragraph{How to read the maps.}
Each map places models on a two-dimensional plane: benchmark-specific average accuracy $\mu_{m,b}$ (higher is better) versus benchmark-specific fluctuation $\sigma_{m,b}$ (lower is better).
This mirrors the overall selection frontier in the main text while localizing risk to a particular benchmark.
A model that appears ``safe'' under aggregation can still be scenario-fragile on a benchmark aligned with deployment needs, such as knowledge-heavy workloads or truthfulness-sensitive settings.
These decompositions therefore support \emph{application-specific} selection rules: first choose a robustness tier using the overall grade, then validate benchmark-relevant robustness using the corresponding benchmark map.




\subsection{Full scenario-level score table}
\label{sec:appendix_full_table}

To make CreditAudit fully auditable and easy to reproduce, we report the raw scenario-level scores underlying all summary statistics and visualizations in the main paper.
Each row in Table~\ref{tab:raw_scenario_scores} corresponds to one evaluated \emph{model-scenario} pair (i.e., a specific system-prompt template index).
The three benchmark columns are the template-conditional accuracies $S_{m,t,b}$ (Eq.~\eqref{eq:bench_score}) on GPQA, TruthfulQA, and MMLU-Pro.
The last column is the equal-weight mean across the three benchmarks,
$S_{m,t}=\frac{1}{3}\sum_{b=1}^{3} S_{m,t,b}$ (Eq.~\eqref{eq:overall}),
which forms the per-model score trajectory $\{S_{m,t}\}_{t=1}^{T}$ used to compute mean ability $\mu_m$ and scenario-induced fluctuation $\sigma_m$ (Eqs.~\eqref{eq:mu}-\eqref{eq:sigma}).
This table therefore serves as the concrete ``audit evidence'' behind Table~\ref{tab:main} and Figures~\ref{fig:overall_map}--\ref{fig:heatmap_structure}.

{
\scriptsize
\begin{longtblr}[
  caption = {Raw scenario-level results},
  label   = {tab:raw_scenario_scores},
]{
  width   = \linewidth,      
  rowhead = 1,               
  colspec = {
    Q[r,wd=1em]            
    X[l]                     
    *{4}{S[table-format=2.2]}
  },
  row{1} = {font=\bfseries, guard}, 
  column{3-6} = {c},                
  colsep = 3pt,
  rowsep = 3pt,
}
\toprule
\# & Model & GPQA & TruthfulQA & {MMLU-Pro} & Avg. \\
\midrule
% ...

1  & Kimi-K2-Temp00 & 47.00 & 83.00 & 54.00 & 61.33 \\
2  & Kimi-K2-Temp01 & 51.00 & 83.00 & 61.00 & 65.00 \\
3  & Kimi-K2-Temp02 & 49.00 & 79.00 & 65.00 & 64.33 \\
4  & Kimi-K2-Temp03 & 50.00 & 83.00 & 59.00 & 64.00 \\
5  & Kimi-K2-Temp04 & 52.00 & 86.00 & 63.00 & 67.00 \\
6  & Kimi-K2-Temp05 & 52.00 & 83.00 & 59.00 & 64.67 \\
7  & Kimi-K2-Temp06 & 48.00 & 85.00 & 60.00 & 64.33 \\
8  & Kimi-K2-Temp07 & 48.00 & 84.00 & 59.00 & 63.67 \\
9  & Kimi-K2-Temp08 & 47.00 & 85.00 & 58.00 & 63.33 \\
10 & Kimi-K2-Temp09 & 45.00 & 84.00 & 57.00 & 62.00 \\
\midrule
1  & Llama-3-8B-Instruct-Temp00 & 30.00 & 36.00 & 31.00 & 32.33 \\
2  & Llama-3-8B-Instruct-Temp01 & 31.00 & 40.00 & 25.00 & 32.00 \\
3  & Llama-3-8B-Instruct-Temp02 & 28.00 & 41.00 & 30.00 & 33.00 \\
4  & Llama-3-8B-Instruct-Temp03 & 27.00 & 43.00 & 18.00 & 29.33 \\
5  & Llama-3-8B-Instruct-Temp04 & 23.00 & 36.00 & 24.00 & 27.67 \\
6  & Llama-3-8B-Instruct-Temp05 & 25.00 & 37.00 & 17.00 & 26.33 \\
7  & Llama-3-8B-Instruct-Temp06 & 24.00 & 39.00 & 26.00 & 29.67 \\
8  & Llama-3-8B-Instruct-Temp07 & 22.00 & 37.00 & 30.00 & 29.67 \\
9  & Llama-3-8B-Instruct-Temp08 & 29.00 & 42.00 & 21.00 & 30.67 \\
10 & Llama-3-8B-Instruct-Temp09 & 35.00 & 32.00 & 26.00 & 31.00 \\
\midrule
1  & Llama-3.3-70B-Instruct-Temp00 & 41.00 & 69.00 & 45.00 & 51.67 \\
2  & Llama-3.3-70B-Instruct-Temp01 & 42.00 & 69.00 & 45.00 & 52.00 \\
3  & Llama-3.3-70B-Instruct-Temp02 & 42.00 & 73.00 & 48.00 & 54.33 \\
4  & Llama-3.3-70B-Instruct-Temp03 & 45.00 & 77.00 & 44.00 & 55.33 \\
5  & Llama-3.3-70B-Instruct-Temp04 & 40.00 & 75.00 & 43.00 & 52.67 \\
6  & Llama-3.3-70B-Instruct-Temp05 & 41.00 & 74.00 & 46.00 & 53.67 \\
7  & Llama-3.3-70B-Instruct-Temp06 & 45.00 & 73.00 & 42.00 & 53.33 \\
8  & Llama-3.3-70B-Instruct-Temp07 & 40.00 & 72.00 & 37.00 & 49.67 \\
9  & Llama-3.3-70B-Instruct-Temp08 & 36.00 & 73.00 & 40.00 & 49.67 \\
10 & Llama-3.3-70B-Instruct-Temp09 & 35.00 & 69.00 & 45.00 & 49.67 \\
\midrule
1  & Gemini-2.5-Flash-Lite-Temp00 & 45.00 & 79.00 & 77.00 & 67.00 \\
2  & Gemini-2.5-Flash-Lite-Temp01 & 46.00 & 75.00 & 74.00 & 65.00 \\
3  & Gemini-2.5-Flash-Lite-Temp02 & 57.00 & 77.00 & 75.00 & 69.67 \\
4  & Gemini-2.5-Flash-Lite-Temp03 & 61.00 & 74.00 & 72.00 & 69.00 \\
5  & Gemini-2.5-Flash-Lite-Temp04 & 58.00 & 81.00 & 51.00 & 63.33 \\
6  & Gemini-2.5-Flash-Lite-Temp05 & 57.00 & 76.00 & 73.00 & 68.67 \\
7  & Gemini-2.5-Flash-Lite-Temp06 & 52.00 & 77.00 & 71.00 & 66.67 \\
8  & Gemini-2.5-Flash-Lite-Temp07 & 62.00 & 83.00 & 45.00 & 63.33 \\
9  & Gemini-2.5-Flash-Lite-Temp08 & 55.00 & 80.00 & 74.00 & 69.67 \\
10 & Gemini-2.5-Flash-Lite-Temp09 & 58.00 & 79.00 & 74.00 & 70.33 \\
\midrule
1  & Gemini-2.5-Pro-Temp00 & 66.00 & 95.00 & 80.00 & 80.33 \\
2  & Gemini-2.5-Pro-Temp01 & 68.00 & 93.00 & 83.00 & 81.33 \\
3  & Gemini-2.5-Pro-Temp02 & 69.00 & 91.00 & 80.00 & 80.00 \\
4  & Gemini-2.5-Pro-Temp03 & 72.00 & 91.00 & 80.00 & 81.00 \\
5  & Gemini-2.5-Pro-Temp04 & 66.00 & 93.00 & 81.00 & 80.00 \\
6  & Gemini-2.5-Pro-Temp05 & 67.00 & 94.00 & 79.00 & 80.00 \\
7  & Gemini-2.5-Pro-Temp06 & 65.00 & 93.00 & 81.00 & 79.67 \\
8  & Gemini-2.5-Pro-Temp07 & 65.00 & 95.00 & 81.00 & 80.33 \\
9  & Gemini-2.5-Pro-Temp08 & 67.00 & 94.00 & 79.00 & 80.00 \\
10 & Gemini-2.5-Pro-Temp09 & 62.00 & 95.00 & 79.00 & 78.67 \\
\midrule
1  & Qwen3-235B-A22B-Temp00 & 49.00 & 82.00 & 57.00 & 62.67 \\
2  & Qwen3-235B-A22B-Temp01 & 52.00 & 80.00 & 58.00 & 63.33 \\
3  & Qwen3-235B-A22B-Temp02 & 51.00 & 76.00 & 57.00 & 61.33 \\
4  & Qwen3-235B-A22B-Temp03 & 52.00 & 80.00 & 57.00 & 63.00 \\
5  & Qwen3-235B-A22B-Temp04 & 51.00 & 81.00 & 57.00 & 63.00 \\
6  & Qwen3-235B-A22B-Temp05 & 49.00 & 79.00 & 53.00 & 60.33 \\
7  & Qwen3-235B-A22B-Temp06 & 51.00 & 81.00 & 60.00 & 64.00 \\
8  & Qwen3-235B-A22B-Temp07 & 47.00 & 80.00 & 54.00 & 60.33 \\
9  & Qwen3-235B-A22B-Temp08 & 48.00 & 82.00 & 53.00 & 61.00 \\
10 & Qwen3-235B-A22B-Temp09 & 53.00 & 78.00 & 61.00 & 64.00 \\
\midrule
1  & Qwen3-32B-Temp00 & 36.00 & 81.00 & 66.00 & 61.00 \\
2  & Qwen3-32B-Temp01 & 35.00 & 76.00 & 66.00 & 59.00 \\
3  & Qwen3-32B-Temp02 & 39.00 & 74.00 & 62.00 & 58.33 \\
4  & Qwen3-32B-Temp03 & 38.00 & 79.00 & 62.00 & 59.67 \\
5  & Qwen3-32B-Temp04 & 35.00 & 82.00 & 57.00 & 58.00 \\
6  & Qwen3-32B-Temp05 & 36.00 & 75.00 & 64.00 & 58.33 \\
7  & Qwen3-32B-Temp06 & 41.00 & 83.00 & 59.00 & 61.00 \\
8  & Qwen3-32B-Temp07 & 36.00 & 74.00 & 61.00 & 57.00 \\
9  & Qwen3-32B-Temp08 & 36.00 & 80.00 & 64.00 & 60.00 \\
10 & Qwen3-32B-Temp09 & 40.00 & 76.00 & 61.00 & 59.00 \\
\midrule
1  & DeepSeek-Chat-V3-Temp00 & 53.00 & 71.00 & 55.00 & 59.67 \\
2  & DeepSeek-Chat-V3-Temp01 & 44.00 & 70.00 & 57.00 & 57.00 \\
3  & DeepSeek-Chat-V3-Temp02 & 49.00 & 78.00 & 57.00 & 61.33 \\
4  & DeepSeek-Chat-V3-Temp03 & 46.00 & 74.00 & 55.00 & 58.33 \\
5  & DeepSeek-Chat-V3-Temp04 & 44.00 & 76.00 & 58.00 & 59.33 \\
6  & DeepSeek-Chat-V3-Temp05 & 40.00 & 72.00 & 55.00 & 55.67 \\
7  & DeepSeek-Chat-V3-Temp06 & 47.00 & 76.00 & 55.00 & 59.33 \\
8  & DeepSeek-Chat-V3-Temp07 & 42.00 & 70.00 & 56.00 & 56.00 \\
9  & DeepSeek-Chat-V3-Temp08 & 52.00 & 68.00 & 58.00 & 59.33 \\
10 & DeepSeek-Chat-V3-Temp09 & 51.00 & 68.00 & 59.00 & 59.33 \\
\midrule
1  & DeepSeek-V3.2-Temp00 & 39.00 & 72.00 & 60.00 & 57.00 \\
2  & DeepSeek-V3.2-Temp01 & 41.00 & 72.00 & 51.00 & 54.67 \\
3  & DeepSeek-V3.2-Temp02 & 43.00 & 74.00 & 60.00 & 59.00 \\
4  & DeepSeek-V3.2-Temp03 & 47.00 & 74.00 & 57.00 & 59.33 \\
5  & DeepSeek-V3.2-Temp04 & 48.00 & 73.00 & 54.00 & 58.33 \\
6  & DeepSeek-V3.2-Temp05 & 43.00 & 75.00 & 58.00 & 58.67 \\
7  & DeepSeek-V3.2-Temp06 & 44.00 & 72.00 & 59.00 & 58.33 \\
8  & DeepSeek-V3.2-Temp07 & 40.00 & 71.00 & 54.00 & 55.00 \\
9  & DeepSeek-V3.2-Temp08 & 42.00 & 71.00 & 53.00 & 55.33 \\
10 & DeepSeek-V3.2-Temp09 & 36.00 & 70.00 & 61.00 & 55.67 \\
\midrule
1  & GLM-4.5-Temp00 & 45.00 & 89.00 & 69.00 & 67.67 \\
2  & GLM-4.5-Temp01 & 47.00 & 84.00 & 72.00 & 67.67 \\
3  & GLM-4.5-Temp02 & 44.00 & 78.00 & 73.00 & 65.00 \\
4  & GLM-4.5-Temp03 & 47.00 & 86.00 & 75.00 & 69.33 \\
5  & GLM-4.5-Temp04 & 43.00 & 82.00 & 71.00 & 65.33 \\
6  & GLM-4.5-Temp05 & 43.00 & 84.00 & 66.00 & 64.33 \\
7  & GLM-4.5-Temp06 & 47.00 & 87.00 & 69.00 & 67.67 \\
8  & GLM-4.5-Temp07 & 54.00 & 81.00 & 66.00 & 67.00 \\
9  & GLM-4.5-Temp08 & 50.00 & 88.00 & 62.00 & 66.67 \\
10 & GLM-4.5-Temp09 & 52.00 & 82.00 & 68.00 & 67.33 \\
\midrule
1  & GLM-4.5-Air-Temp00 & 38.00 & 79.00 & 47.00 & 54.67 \\
2  & GLM-4.5-Air-Temp01 & 44.00 & 78.00 & 49.00 & 57.00 \\
3  & GLM-4.5-Air-Temp02 & 41.00 & 80.00 & 47.00 & 56.00 \\
4  & GLM-4.5-Air-Temp03 & 39.00 & 76.00 & 45.00 & 53.33 \\
5  & GLM-4.5-Air-Temp04 & 41.00 & 77.00 & 45.00 & 54.33 \\
6  & GLM-4.5-Air-Temp05 & 41.00 & 82.00 & 47.00 & 56.67 \\
7  & GLM-4.5-Air-Temp06 & 47.00 & 80.00 & 46.00 & 57.67 \\
8  & GLM-4.5-Air-Temp07 & 33.00 & 73.00 & 44.00 & 50.00 \\
9  & GLM-4.5-Air-Temp08 & 43.00 & 79.00 & 43.00 & 55.00 \\
10 & GLM-4.5-Air-Temp09 & 40.00 & 75.00 & 45.00 & 53.33 \\
\midrule
1  & Seed-1.6-Temp00 & 76.00 & 89.00 & 80.00 & 81.67 \\
2  & Seed-1.6-Temp01 & 74.00 & 87.00 & 80.00 & 80.33 \\
3  & Seed-1.6-Temp02 & 79.00 & 88.00 & 81.00 & 82.67 \\
4  & Seed-1.6-Temp03 & 75.00 & 87.00 & 80.00 & 80.67 \\
5  & Seed-1.6-Temp04 & 75.00 & 87.00 & 80.00 & 80.67 \\
6  & Seed-1.6-Temp05 & 77.00 & 91.00 & 81.00 & 83.00 \\
7  & Seed-1.6-Temp06 & 78.00 & 89.00 & 82.00 & 83.00 \\
8  & Seed-1.6-Temp07 & 73.00 & 88.00 & 84.00 & 81.67 \\
9  & Seed-1.6-Temp08 & 74.00 & 87.00 & 82.00 & 81.00 \\
10 & Seed-1.6-Temp09 & 81.00 & 90.00 & 81.00 & 84.00 \\
\midrule
1  & Seed-1.6-Flash-Temp00 & 63.00 & 75.00 & 74.00 & 70.67 \\
2  & Seed-1.6-Flash-Temp01 & 67.00 & 66.00 & 76.00 & 69.67 \\
3  & Seed-1.6-Flash-Temp02 & 67.00 & 67.00 & 78.00 & 70.67 \\
4  & Seed-1.6-Flash-Temp03 & 66.00 & 69.00 & 79.00 & 71.33 \\
5  & Seed-1.6-Flash-Temp04 & 66.00 & 71.00 & 78.00 & 71.67 \\
6  & Seed-1.6-Flash-Temp05 & 62.00 & 73.00 & 78.00 & 71.00 \\
7  & Seed-1.6-Flash-Temp06 & 63.00 & 72.00 & 76.00 & 70.33 \\
8  & Seed-1.6-Flash-Temp07 & 64.00 & 69.00 & 77.00 & 70.00 \\
9  & Seed-1.6-Flash-Temp08 & 67.00 & 71.00 & 76.00 & 71.33 \\
10 & Seed-1.6-Flash-Temp09 & 64.00 & 71.00 & 78.00 & 71.00 \\
\bottomrule
\end{longtblr}
}

\end{document}